Article

# Selecting the Suitable Resampling Strategy for Imbalanced Data Classification Regarding Dataset Properties. An Approach Based on Association Models

Mohamed S. Kraiem, Fernando Sánchez-Hernández and María N. Moreno-García *

Data Mining Research Group, University of Salamanca, 37008 Salamanca, Spain; ing_kriem@yahoo.com (M.S.K.); fsh@usal.es (F.S.-H.)
* Correspondence: mmg@usal.es

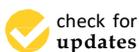

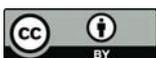

**Abstract:** In many application domains such as medicine, information retrieval, cybersecurity, social media, etc., datasets used for inducing classification models often have an unequal distribution of the instances of each class. This situation, known as imbalanced data classification, causes low predictive performance for the minority class examples. Thus, the prediction model is unreliable although the overall model accuracy can be acceptable. Oversampling and undersampling techniques are well-known strategies to deal with this problem by balancing the number of examples of each class. However, their effectiveness depends on several factors mainly related to data intrinsic characteristics, such as imbalance ratio, dataset size and dimensionality, overlapping between classes or borderline examples. In this work, the impact of these factors is analyzed through a comprehensive comparative study involving 40 datasets from different application areas. The objective is to obtain models for automatic selection of the best resampling strategy for any dataset based on its characteristics. These models allow us to check several factors simultaneously considering a wide range of values since they are induced from very varied datasets that cover a broad spectrum of conditions. This differs from most studies that focus on the individual analysis of the characteristics or cover a small range of values. In addition, the study encompasses both basic and advanced resampling strategies that are evaluated by means of eight different performance metrics, including new measures specifically designed for imbalanced data classification. The general nature of the proposal allows the choice of the most appropriate method regardless of the domain, avoiding the search for special purpose techniques that could be valid for the target data.

## 1. Introduction

Several real-world datasets are characterized by a highly imbalanced distribution of instances through the classes. Then, one or more classes, known as the minority class(es), contain a much smaller number of examples than the other class(es), the majority class(es) [1]. Most classifiers are very sensitive to skewed distributions, providing an outcome mainly based on the majority class(es). Thus, the precision for the minority class(es) is usually significantly lower than the precision for the majority class(es), which is an indication of an unreliable prediction model. Frequently, the minority class(es) is (are) the most interesting regarding the application domain under study, therefore, the problem gets worse in these cases. An example of this situation is cancer diagnosis, where the minority class, corresponding to cancer patients, is the most important. Therefore, this class requires the highest precision, since the cost of misclassifying a cancer patient as false negative is much higher than classifying a healthy patient as false positive [2].

Increasing reliability of imbalanced data classification is a challenge in many areas where this important drawback is very common. Healthcare and medical diagnosis are two of them, in which this problem is particularly relevant. The way to deal with it has been addressed in studies about infections acquired in intensive care units [3,4], indication of noninvasive mechanical ventilation [5] and predicting neurological outcome of patients with severe trauma [6,7] among others. In the field of information retrieval this drawback is also very present, especially in areas such as text classification or sentiment analysis, where relevant documents and sentiment polarity/intensity respectively show a clear class imbalance [8,9]. Web spam classification also requires resampling methods for data preprocessing [10]. Fraud detection related to bank transactions, credit cards, insurance companies, etc. [11], software fault prediction [12] and profile injection attacks in recommender systems [13] are other significant application domains of machine learning methods where data are imbalanced.

To address this problem, many proposals have been made at level of datasets as well as at level of algorithms. These approaches are also known as external and internal methods, respectively. The former aim to mitigate the imbalance by resampling data, while the latter involve the design of specific algorithms for classifying imbalanced data.

Oversampling and undersampling external techniques are the most well-known strategies to deal with this problem by means of balancing the different number of examples of each class. Dataset oversampling involves the replication of minority class examples, while undersampling is the process of removing majority class examples. The effectiveness of these strategies is affected by some factors whose study is one of the main purposes of this work.

Factors affecting efficacy of resampling strategies are mainly associated to data intrinsic properties as imbalance ratio or dataset size, among others [14]. In this work, the impacts of dataset imbalance ratio, number of instances, number of attributes, overlapping between classes and borderline examples are analyzed through a wide comparative study where the classification results are examined to determine which resampling method is more suitable regarding the characteristics of the dataset. In order to cover a wide range of conditions, 40 datasets from different application areas have been tested. These datasets encompass a broad interval of values of the factors under study.

Aiming at achieving a deep analysis of results, eight evaluation metrics have been applied to compare the outcomes of the classifiers. These were induced from imbalanced datasets before and after preprocessing them by means of a total of seven sampling strategies, both basic and advanced. General measures, such as accuracy, as well as specific ones for imbalance problems, such as optimized precision [15] and generalized index of balanced accuracy [16], have been used.

The final objective is to induce association models [17] to assist in the selection of the resampling strategy that is most suitable for any dataset regardless of which field it belongs to, since adequacy is based only on data properties. The great diversity of the datasets involved in the study and the volume of results used to generate the association rules make these models very useful for decision making in a wide variety of conditions.

In fact, the main contribution of this work is the generation of the associative models that involve several factors together, providing the best resampling method for different combinations of values of these factors. As far as we know, no other work in the literature has focused on the joint evaluation of such characteristics or considered such a broad spectrum of their values, especially those of imbalance ratio. In addition, the resampling methods evaluated, and the metrics used exceed in quantity and variety those studied in most related research.

## 2. Related Work

There is an extensive literature focused on this problem, which is very common in many domains. In this section, we include a brief survey of some representative work that addresses the improvement of imbalanced data classification by using both internal and external algorithms.

Solutions at the internal level include some techniques such as modifying existing classification algorithms, one-class classification and cost-sensitive methods. Some internal strategies such as instance weighting can be used either outside this classification to give more weight to minority class instances or within cost-sensitive methods [14]. Another option to reduce the bias toward majority class instances is to calibrate the prediction probability of the algorithms to align it with prior probabilities.

The aim of algorithm modification is to make the resulting classifier more suitable for class imbalance distributions. In order to achieve this objective, a process of learning from these distributions is performed prior to the induction of the classification model. A well-known classifier that has been adapted to process imbalanced datasets is SVM. Some modifications of this algorithm are Z-SVM [18], GSVM and GSVM-RU [19]. Z-SVM aims to increase the G-mean values by moving the position of the hyperplane of the original algorithm. GSVM and GSVM-RU are based on granular computing, consisting of splitting the problem into a sequence of granules and building SVMs on some of these granules. GSVM-RU also uses SVM for undersampling, which makes it suitable for high imbalance problems. Fuzzy classification is another approach proposed for some authors to deal with the problem of imbalanced data. In [20], a fuzzy-based classifier is presented where fuzzy sets representing the classes are created from the relative class frequency distributions. Hierarchical fuzzy rules (HFR) and a genetic algorithm are combined in [21] to classify imbalanced datasets. Some variations of SVM based on fuzzy sets have been proposed to deal with the problem of classification of nonbalanced data [22,23].

A less popular alternative is to resort to one-class classifiers that are focused on classifying instances of a target class, ignoring all other classes [24]. These methods are generally based on similarity measures between examples of the target class. In the class imbalance scenario, this approach has been applied in some works in the literature [25–27] where the minority class is considered the target class and the majority class as the outlier. Although only the minority class is learned, examples from the majority class can also be used for training. This approach is also combined with resampling strategies [28].

Cost-sensitive learning is an internal technique based on considering the misclassification cost, which has been the subject of more studies in the literature. In [29], the impact of imbalance on classifiers is analyzed when the misclassification costs are unequal. The work of [30] includes an effective framework that integrates an evaluation measure into the objective function of cost-sensitive SVM directly to increase the performance by improving misclassification cost criteria. The use of an uneven cost matrix in an approach based on neural networks and fuzzy logic is presented in [31]. One of the problems of cost-sensitive learning is the frequent unavailability of the cost matrix, thus this drawback has also been addressed in some work. Some authors propose to create artificial cost values when cost values cannot be obtained [32]. The problem of overfitting is also analyzed in the literature on these methods [33], concluding that this drawback associated with oversampling strategies also occurs with cost-sensitive methods. On the other hand, both approaches present similar performance when dealing with the class imbalance problem [34].

Regarding external methods, data resampling strategies have received considerable attention in machine learning since they are one of the most reliable approaches to deal with the class imbalance problem. Resampling methods try to overcome imbalanced class distribution drawback by either adding or removing examples from the dataset.

Among all the sampling strategies in the literature, SMOTE (Synthetic Minority Over-Sampling Technique) [35] is one of the most widely used. In this method, a nearest neighbor-based interpolation procedure is used to generate examples of the minority class. Tomek Link (T-Link) [36] is other popular technique that also considers nearest neighbor examples, but in this case for undersampling purposes. Other commonly used undersampling, oversampling and hybrid approaches are described in detail in the next section. They have been proposed aiming at improving the precision of the minority class in imbalanced data contexts. However, their behavior is not the same for all datasets.

There are also hybrid approaches that combine specific algorithms with resampling techniques. A fuzzy rule-based classifier that is used with resampling techniques in [37]. In [38] a new data reduction procedure is presented to be used in combination with an SVM-based classification algorithm. The method, based on the fuzzy adaptive resonance theory, involves the modification of the data distribution.

According to [39], classifier performance depends on some factors such as imbalance ratio between the majority and minority classes, dataset characteristics, classification algorithms and resampling methods. Classification model reliability can be significantly improved when imbalance ratio between the majority and the minority class is reduced by using suitable resampling methods. The effects of the mentioned factors on the performance have been evaluated in some works, however, either factors are mainly analyzed in an individual way without considering them jointly or a not very wide range of data property values are considered. Moreover, in most of the works, classic metrics are used for validation.

In [40], the effect of basic resampling methods on the performance of C5.0 decision trees is discussed. Classifiers induced from a reduced number of real-world datasets are tested by obtaining the error rate on each class. A study of the effectiveness of several sampling techniques is presented in [41]. In this work, improvement of the performance of a logistic regression classifier in presence of data imbalance is examined aiming at finding out the best strategy to be used. Only the precision of the majority class and the recall of the minority class are obtained for evaluating the effect of the sampling strategies on the classification. Other drawback of this work is the use of a unique synthetic dataset in the study. Although there are some other studies about resampling techniques, their authors do not specify which resampling techniques can be used according to specific datasets characteristics and machine learning algorithms.

The behavior of ensemble classifiers with imbalanced data has also been a subject under study. Diez-Pastor et al. [42] presented an extensive experimental study where various ensemble-based methods are tested in its original form and modified with several diversity enhancing techniques. All the algorithms are applied to 84 imbalanced datasets before and after preprocessing them with well-known undersampling and oversampling techniques. J48 (Java implementation of C4.5) is the base classifier used in all ensembles. The goal of the authors is to show that techniques for diversity increasing improve the performance of the ensemble methods for imbalance problems, but this does not always occur. Thus, metafeatures characterizing the complexity of the datasets are used in this study to know when its application is appropriate. However, features such as imbalance ratio are not considered. On the other hand, validation is carried out by means of three classic metrics (F-measure, G-mean and AUC), which sometimes give contradictory results. No advanced metrics for imbalanced data classification have been applied.

Other work where the behavior of base and ensemble classifiers is analyzed in imbalanced data contexts was carried out in [14]. The authors performed a review of the methods proposed for dealing with the problem of imbalanced data classification and provided a taxonomy to classify them. The goal of the work is to analyze the preprocessing methods, including resampling strategies, and compare their efficacy as a step prior to the classification. The cost of the incorrect classifications for each class is also considered, which requires involving human experts. Some data intrinsic characteristics are analyzed in the study. The authors conclude that issues such as the lack of density, small sample size and class overlapping have a greater influence in the performance of classifiers than the imbalance ratio. However, these conclusions cannot be generalized since the datasets used in the study have a low imbalance ratio and most of them have a low number of attributes and examples. In [4], a clustering-based undersampling strategy combined with ensemble classifiers is proposed and tested with only one dataset in a medical application domain. An ensemble approach is also presented in [43] for assisted reproductive technology outcome prediction. The ensemble is built with two incremental classifiers that can update themselves with new data. Both studies focus on solving the imbalance problem in very specific cases.

The impact of class imbalance ratio for a particular kind of classifiers has also been analyzed in other studies whose main purpose is helping to select the best resampling method regarding the values of that factor [44]. Other works focused on specific classification algorithms are devoted to proposing sampling approaches to improve their performance. In [45], a hybrid procedure combining both oversampling and undersampling with a support vector machine (SVM) ensemble manages to improve the reliability of that classifier. A random forests quantile classifier is proposed in [46] for class imbalanced data. Other study addresses imbalanced data classification in the area of cancer diagnosis by means of a feature selection method and a cost sensitive support vector machine (CSSVM) learning algorithm [2].

The study of the performance of neural networks and deep learning algorithms in unbalanced data contexts is much scarcer, being mostly focused in the areas of image and text processing where the volume of data is very high. Generative adversarial networks (GAN) have been used as basis of oversampling strategies in [47]. There are proposals in the literature for using them to generate sentences in text processing [10] and variants of GANT to generate synthetic images [48] in order to balance data. Complexity and computational cost are the main drawbacks of this approach. In addition, text generation models for imbalanced datasets based on deep learning algorithms, such as LSTM (long short-term memory) networks, are also found in the literature [49].

The performance of deep learning algorithms on imbalanced datasets is similar to that of traditional algorithms, except when working with large datasets. For this reason, these techniques are also often used in combination with resampling strategies in the presence of uneven distributions [50,51]. In [50], several classifiers were tested on a highly imbalanced dataset composed by 146,500 instances, both in its original form and resampled by several undersampling and oversampling strategies. The results showed that random forest (RF) outperformed deep learning algorithms. Two imbalanced datasets of similar size to the above reference were used in [51] to test a resampling technique proposed in the work with different machine learning algorithms including RF and LSTM. The proposal was compared with popular resampling techniques such as random undersampling (RUS), random oversampling (ROS) or SMOTE. In most experiments RF provided the highest values for all performance metrics. The results of LSTM were unequal since only with RUS and for one of the datasets LSTM achieved the best values for all metrics. From the analysis of these and other works we can deduce that random forest combined with different resampling strategies can be a good choice to deal with classification of unbalanced data. Only in domains with high data volume, such as image or text processing, may the use of deep learning techniques be more appropriate.

The focus of some proposals in the literature is improving the classification performance in specific application domains. In [52], SVM is used jointly with SMOTE oversampling technique for preprocessing imbalanced data related to caravan car policy. This study analyzes the values of sensitivity obtained after applying several machine learning algorithms to preprocessed data. Given that the work is focused on a unique dataset, factors affecting performance of the prediction are not studied. Moreover, no recent specific metrics for imbalanced data classification are used in the validation. In the field of sentiment classification, there are a number of similar proposals, such as the modification of a SVM-based ensemble algorithm by incorporating both oversampling and undersampling [53]. In the area of vaccine design, the method CluSMOTE [54], which combines a cluster-based undersampling approach and a synthetic minority oversampling technique, is proposed to improve the results of AUC and G-mean provided by the SVM algorithm. Web application security is a field where the imbalanced data problem is also present. Mokbal et al. [55] tackled it by means of a novel approach for generating synthetic samples of minority class that have identical distribution as real scenarios by using generative adversarial networks. All these proposals are addressed to specific classification methods or domains, making them difficult to apply across the board.

Other research interest is to improve existing resampling methods. In [56], the SMOTE strategy is extended by adding an iterative ensemble-based noise filter called iterative partitioning filter (IPF) in order to deal with problems in the classification performance produced by noisy and borderline examples in imbalanced datasets. The results of this extension, named SMOTE-IPF, are better than those yielded by the basic SMOTE and other SMOTE-based methods. The validation is carried out with different real and synthetic datasets in a not too large range of number of instances, number of attributes and imbalance ratio. Cateni and coworkers [57] proposed a new resampling method by combining oversampling and undersampling techniques. In this study the authors try to minimize both the overfitting problem that happens when using random oversampling and the information lost associated to the random undersampling technique. Their results cannot be generalized since the method has not been tested with highly imbalanced datasets. In [58], SMOTE is combined with undersampling strategies such as edited nearest neighbor or Tomek link, explained in the nest sections, to address the multiclass imbalance problem. Zeraatkar et al. [59] present a proposal for improvement of SMOTE and its variants based on oversampling and undersampling. SMOTE is used in the first step for oversampling. Then, the undersampling process is performed by new extensions of KNN classifiers based on fuzzy sets. Recently, deep learning-based oversampling methods such as variational autoencoders have been developed [60,61] although they are mainly indicated for image processing.

The ultimate goal of this study is to find patterns that relate the characteristics of the data to the most appropriate resampling strategy according to those characteristics. Association rules are a valuable tool for finding patterns, especially in contexts of data sparsity. Moreover, these patterns are usually less complex and more interpretable than those obtained with supervised machine learning algorithms [62]. There is a wide variety of association rule algorithms in the literature, ranging from the Apriori algorithm [63] to sophisticated proposals for improving it [64]. Most of them, such as the FP-Grown algorithm [65], are focused on reducing the computational cost, which is very high in large datasets. In fact, specific algorithms have been designed for large databases [66]. Another important goal of some algorithms is to reduce the number of rules to facilitate the interpretation of patterns [67,68]. However, none of these drawbacks are present with the data used to generate association rules in this work.

The way of validating the performance of classifiers has been the subject of some studies. The main metrics that have been proposed for unbalanced data classification are described in the next section.

In this paper we present a novel contribution addressed to analyze the effectiveness of different resampling techniques to improve classifier outcome depending on some dataset characteristics. The objective is to provide a systematic way of selecting the most suitable strategy for every dataset under study.

## 3. Materials and Methods

In this section we describe the resampling strategies, the data characteristics analyzed, and the metrics used to evaluate the results.

### 3.1. Resampling Strategies

The usual way of dealing with the problem of imbalanced datasets is to resort to resampling techniques in order to achieve a more balanced distribution of instances of the classes. There are two kinds of basic resampling strategies that can be used, random undersampling and random oversampling. These basic approaches present some drawbacks that have been addressed by means of hybrid strategies and advanced methods. All these techniques are discussed below.

Random undersampling (RUS) methods remove examples from the majority class to balance the dataset. The drawback of random undersampling is the loss of potentially useful information that could be important for classification, causing a low performance of the model. This method can be suitable for slightly imbalanced datasets, but it is not recommended for highly imbalanced data.

Random oversampling (ROS) techniques, also called upsampling, deal with the dataset imbalance by means of replicating the examples of the minority class. In these methods there are no loss of information, but they present the disadvantage of overfitting as well as the introduction of an additional computational cost if the imbalance ratio is high. ROS is efficient for highly imbalanced datasets and its use generally leads to high performance models.

Hybrid sampling strategies combine oversampling and undersampling in different ways. However, drawbacks of both sampling techniques can also be present in this approach.

### 3.1.1. SMOTE

The main drawbacks of random oversampling have been addressed by introducing some modifications in the original algorithm and evolving to more advanced approaches. One of the most popular is SMOTE (synthetic minority oversampling technique) [35], an oversampling method proposed for dealing with the overfitting problem, making the decision border of the minority class more general. In this method, instead of replicating instances of the minority class, new examples are generated by interpolating the existing ones. The steps of the algorithm are the following:

1. Randomly select *k* nearest neighbors for each training instance *x* of the minority class.
2. Select a random point along the line segment between the instance *x* and the nearest neighbor and add it as a new instance to the training set.

SMOTE is an algorithm especially efficient for highly imbalanced datasets and its performance can be improved combining it with undersampling techniques [35].

### 3.1.2. Condensed Nearest Neighbor

Condensed nearest neighbor (CNN) is a strategy initially proposed by Hart [69] as a data reduction method used to improve the efficiency in the implementation of the k nearest neighbor decision rule (k-NN) for classification problems. In the context of data reduction methods when k-NN is used as classifier, the examples in the training set fall into one of three categories:

- Outliers: points whose k nearest neighbors do not belong to the same class. That is, examples that are incorrectly classified by k-NN.
- Prototypes: the minimum set of points of the training set needed to classify correctly nonoutlier points.
- Absorbed points: points that would be correctly classified from the set of prototypes.

CNN uses the basic approach of the NN rule but does not meet all its constraints. This method selects a consistent subset of prototype points from the original dataset, which will be used to classify new instances, but not necessary the minimum subset. The algorithm works as described below:

1. *D* denotes the original dataset and *E* the resulting condensed set containing prototypes.
2. An arbitrary point from *D* is selected and placed in an original empty set *E*.
3. The remaining points in *D* are classified by the 1-NN rule (*k* = 1) using E, and those that are classified incorrectly are added to *E*.
4. This procedure is iterated until no more data points are transferred from *D* to *E*.

### 3.1.3. Edited Nearest Neighbor

Tomek [36] and Wilson [70] proposed similar modifications of CNN. They suggested to preprocess the data before applying CNN with the aim of achieving a significant data reduction while maintaining a low classification error rate [71]. Edited nearest neighbor (ENN) [70] was the proposal of Wilson for undersampling of the majority class by retaining only the instances that are correctly classified by the k-NN rule. On the other hand, according to Tomek, the CNN rule keeps too many points that are not near the decision boundary because of its arbitrary initialization step. In order to address this problem, he proposes a preliminary pass to select a special subset of $D$ called $C$. Then, his method works in the same manner as CNN rule but, instead of moving to $E$ data points from the complete $D$, only data points from $C$ are used.

### 3.1.4. Tomek Link

Tomek link (T-Link) [36] is another advanced sampling technique proposed to enhance the nearest neighbor classification method by means of removing noise and examples in the border of the classes. Although the objective of the method is not returning balanced data, it can be considered as a guided undersampling method when the training instances from the majority class are removed. The basis of the strategy is given below:

1. Two training instances $x$ and $y$ from different classes are considered.
2. The distance between these two instances is denoted by $d(x, y)$.
3. The pair of values $(x, y)$ is called T link if there is not an instance $z$ that fulfils the condition $d(x, z) < d(x, y)$ or $d(y, z) < d(x, y)$.
4. If any two instances are a T-link, either one of them is as a noise or both instances are in the boundary of the classes.

### 3.1.5. One-Sided Selection

One Sided Selection (OSS) strategy is an undersampling technique proposed by Kubat and Matwin [72], which removes instances of the majority class that are considered redundant. The procedure is the following:

1. $D$ is the original dataset.
2. $P$ is a subset of $D$ containing initially all the examples of the minority class from $D$ and one randomly selected example $x$ of the majority class.
3. Using $P$ as training set, the k-nearest neighbors rule (k-NN) with $k = 1$ is used to classify the remaining instances of the majority class.
4. All misclassified examples are moved into the training set $P$, which is smaller than $D$. Correctly classified instances of the majority class are discarded since they are considered redundant.
5. Finally, T-link is used for data cleaning. All borderline and/or noisy examples from the majority class are removed from $P$, while all examples of the minority class are retained.

OSS is indicated for low imbalanced datasets, but it can be used combined with other algorithms to process highly imbalanced datasets.

### 3.2. Data Characteristics Influencing Algorithm Performance

Classification from imbalanced data is a complex problem since it is affected by many factors such as the ratio of imbalance, the overlap between the majority and minority class, borderline examples and small sample size, which causes low performance of the classification models. Detecting these factors in the skewed data and discovering their influence on the classification performance is a research challenge of great interest due to its presence in many areas.

The imbalance ratio (IR) or imbalanced class distribution is defined as the ratio of the number of examples in majority class to the number of examples of the minority class [73]. The imbalance ratio takes place in many cases in real-world datasets and many application domains. Low imbalance ratio and balanced distribution between the classes always gives a better result in the performance of the models. There are different opinions about the effect of class distribution (imbalance ratio), but most of researchers have confirmed that the imbalance ratio between the majority and the minority class is not a problem in itself because there is a relationship between imbalance ratio, borderline examples and overlapping of classes.

Overlapping between classes is an important factor that impacts the performance of the model, especially given the imbalanced distribution of classes. Overlapping means that the minority class instances have properties that are similar to the majority class instances. The overlapping problem makes it difficult for classifiers to identify the limits of a good decision between classes, resulting in decreased performance. The presence of the overlap problem in data with imbalanced distributions becomes increasingly complex because of the need for strategies to balance the data. The application of these strategies, such as oversampling methods, involves the replication of the number of instances in the minority class. This leads to increasing similarity between instances, especially when the imbalance ratio is high and high replication rate is required.

The problem of borderline examples also impacts on the reduction of the classifier efficiency [74]. It means that there are examples in the area around the separation boundaries, where the minority and majority class overlap [75]. The problem of borderline examples is related to noise data. The percentage of borderline examples increases as the number of instances increases because the greater number of instances in the dataset the more noise. This may happen when applying oversampling techniques, such as ROS or SMOTE, because these strategies replicate data, especially in high-imbalance distributions (high imbalance ratio).

Another problem that arises in classification is the small sample size of the training dataset. This factor is related to information lack or density lack. In this case, the algorithm does not have enough information to construct and validate the model. The situation becomes more complicated with imbalance distributions. One of the reasons that cause a small sampling problem in data is the application of random undersampling (RUS) to balance the data, especially when the volume of data is low with a high imbalance ratio, in this case a high percentage of sampling is lost. In addition, the performance of the model is affected when the dataset has a small size and at the same time the number of instances in the minority class is very low. Such a scenario makes the classifier is biased to the majority class; thus, it fails to properly classify the minority class instances and lead to performance loss.

### 3.3. Validation Metrics

Well known metrics as accuracy, precision, recall, F-measure and AUC (area under the ROC curve) [76], were used to validate the classifiers in this work. In addition, other metrics indicated to evaluate models induced from imbalanced data were used: G-mean, optimal precision (OP) and generalized index of balanced accuracy (IBA).

G-mean is the geometric mean of TPR (true positive rate) and TNR (true negative rate), thus, this metric considers the correct classification of instances of both positive and negative classes.

A drawback of AUC and G-mean is the fact that they do not differentiate the contribution of each class to the overall accuracy. Ranawana and Palade [15] addressed this problem and proposed a new metric called optimized precision (OP).

$$OP = \text{Accuracy} - |TNR - TPR| / (TNR + TPR) \tag{1}$$

OP takes the optimal value for a given overall accuracy when true negative rate and true positive rate are very close to each other.

More recently, a new performance measure called generalized index of balanced accuracy (IBA) [16], has been proposed. It can be used and defined for any performance metric (m) as:

$$IBA\alpha(m) = (1 + \alpha \cdot Dom) \cdot m \tag{2}$$

where *Dom*, called dominance, encloses the degree of prevalence of the dominant class over the other. It is defined as:

$$(Dom = TPR - TNR) \tag{3}$$

Given that the values of TPR and TNR are in the range [0, +1], Dom values are within the range [−1, +1]. Values of Dom close to zero, obtained from similar values of true positive rate and true negative rate, indicate that these rates are balanced. In general, the value of Dom is weighted by α (α ≥ 0) to reduce its influence on the result of the metric used (m). In practice, a suitable value of α for a given performance metric must be selected since its value is sensitive to m [77] and affects the accurate of overall IBA result.

## 4. Experimental Study

A wide comparative study has been conducted, in which 40 different datasets have been preprocessed with seven resampling methods. The aim of the study was to determine the factors influencing classifier performance for each sampling strategy after analyzing the results obtained from all metrics referred in the previous section. Original imbalanced datasets were treated in order to reduce their imbalance ratio by means of the following resampling strategies: RUS, ROS, SMOTE and SMOTE combined with OSS, CNN, ENN and TL (T-Link). The percentage used for over sampling was 500% and for undersampling (RUS) 50%. The choice of these values was made after performing some previous experiments and taking into account different criteria. On the one hand, these rates should not be too high to avoid overfitting in the case of oversampling and loss of information in the case of undersampling, and on the other hand, to ensure that the resampling is effective for classification purposes. Working with different values of these rates would have resulted in an unmanageable volume of results due to the number of datasets, resampling methods, evaluation metrics and variables involved in the study. Resampling was performed by using the R unbalanced package (https://cran.r-project.org/web/packages/unbalanced/, last accessed: 6 September 2021), setting k = 6 for SMOTE, k = 1 for CNN and K = 3 for ENN.Random forest (RF) [78] was the machine learning algorithm chosen for inducing the models to be analyzed. We have included only this algorithm in the study since it provided the best results in a previous work [79] where other methods (k-nearest neighbor, naïve Bayes, support vector machine, random forest, J48 and C5.0) were applied for imbalanced data classification. Besides, the behavior of all of them, with the exception of naïve Bayes, was very similar regarding increasing/decreasing of quality metric values for the sampling methods tested in the work. Since the study and report of results of all the algorithms would give an article with an excessive extension, random forest was used to induce predictive models from both original and resampled datasets. Specifically, we make use of the RF implementation given in the random forest package by Breiman and Cutler (https://cran.r-project.org/web/packages/randomForest, last accessed: 6 September 2021) with mtry tuning parameter. It was done through the caret R package (https://github.com/topepo/caret/, last accessed: 6 September 2021). The ntree parameter of the train function was set to 100 and tuneLength parameter for automatic tuning of RF hyperparameters was set to 7.

To calculate the percentages of overlapping and borderline examples, we applied the metrics defined in [80,81] and implemented in the ECoL Library in R (https://github.com/lpfgarcia/ECoL/, last accessed: 6 September 2021). For overlapping, the maximum Fisher's discriminant ratio is used, which measures the overlap between the values of the features and takes the value of the largest discriminant ratio among all the available features. The borderline measure is the fraction of points that are either on the border or in overlapping areas between the classes. It is necessary to build a minimum spanning tree (MST) from data to compute this metric [81].

The validation of the models was performed by means of the metrics included in the previous section. Some of them as accuracy or precision are widely used to evaluate classifiers in any condition, but others such as recall or G-mean are more appropriate for imbalanced data classification. In addition, in the last group, we have included optimized precision (OP) and index of balanced accuracy (IBA) that are even more specific for this kind of problems. IBA was used with α = 0.1 and G-mean as base metric. Both OP and IBA consider the precisions of the majority and minority classes so that a high accuracy of the majority class(es) does not mask the low accuracy of the minority class(es). However, in problems where the latter has a particular relevance, it would be useful to analyze the results for the minority class(es) as well. This particularity has not been taken into account in this study, thus, the selection of the best resampling strategies has been made taking into account the metrics indicated above.

Statistical significance tests were performed to evaluate the differences between the results obtained in the study from the different resampling strategies. In all accomplished experiments to validate the performance of the classifiers, training sets and test sets were generated by using stratified tenfold cross validation with five repetitions. An important aspect of the validation is the fact that resampling strategies were applied only to the folds containing the training sets in each iteration while the test sets were not resampled. The objective is to assure that there are not overfitting problems and the induced classifiers providing good metric values can be applied to real examples different from the training set. When the test sets are formed from the resampled examples, results are usually better, but these results will probably not be reproduced when classifiers are used in a real context.

## 5. Datasets

There is no complete agreement between different authors about when a dataset is considered imbalanced. In this paper, we consider that a dataset is imbalanced when the imbalance ratio (IR) is higher than 1.5, since previous experiments [4,79] have shown that above this threshold, the classification of examples of the minority class(es) is usually significantly lower than the examples of the majority class(es). In addition, many studies in the literature on imbalanced classification start from this IR value to select the datasets [14]. Therefore, this is the lower threshold of the selected datasets. A great variety of datasets are used in this study in order to achieve the objectives described in preceding sections. In order to cover a wide range of imbalance ratio and other data characteristics, the datasets have been selected from different sources:

- Twenty-two datasets taken from KEEL dataset repository.
- Two datasets were taken from the UCI data repository.
- Fifteen datasets created using the datasets.make_classification function of the python Scikit-learn library (GD datasets).
- One dataset (infection) containing anonymous clinical information from patients hospitalized in an intensive care unit.

The following characteristics of each dataset are summarized in Table 1: number of instances (#inst), number of attributes (#attrib), imbalance ratio (IR), percentage of borderline examples (BL%) and percentage of overlapping between the majority and the minority class (OVL%). The datasets are sorted in ascending order according to the imbalance ratio (IR). The datasets included in the study have very varied characteristics, allowing an exhaustive analysis of their impact on the effectiveness of the resampling strategies.

**Table 1.** Information about the datasets used in the study.

| ID | Name of Dataset | #inst | #attrib | IR | BL% | OVL% | Source |
|---|---|---|---|---|---|---|---|
| 1 | Newthyroid | 215 | 5 | 4.84 | 4 | 45 | KEEL |
| 2 | Glass6 | 214 | 9 | 6.38 | 7 | 73 | KEEL |
| 3 | SUSY | 15,000 | 18 | 7.69 | 30 | 87 | UCI |
| 4 | GD2 | 6000 | 6 | 7.74 | 23 | 69 | Scikit learn |
| 5 | Yeast3 | 1484 | 8 | 8.10 | 10 | 63 | KEEL |
| 6 | Ecoli3 | 336 | 7 | 8.6 | 14 | 45 | KEEL |
| 7 | Page-blocks0 | 5472 | 10 | 8.78 | 8 | 19 | KEEL |

Table 1. *Cont.*

| ID | Name of Dataset | #inst | #attrib | IR | BL% | OVL% | Source |
|---|---|---|---|---|---|---|---|
| 8 | Hill_Valley_with_Noise_Testing | 606 | 100 | 8.93 | 24 | 55 | UCI |
| 9 | Ecoli-0-4-6_vs_5 | 203 | 6 | 9.15 | 7 | 49 | KEEL |
| 10 | GD15 | 11,000 | 40 | 9.36 | 15 | 44 | Scikit learn |
| 11 | Vowel0 | 988 | 13 | 9.97 | 2 | 68 | KEEL |
| 12 | Glass-0-1-6_vs_2 | 192 | 9 | 10.29 | 18 | 71 | KEEL |
| 13 | Glass2 | 214 | 9 | 11.59 | 18 | 71 | KEEL |
| 14 | Infection | 4615 | 15 | 13.89 | 14 | 61 | ICU |
| 15 | Ecoli4 | 336 | 7 | 15.80 | 3 | 56 | KEEL |
| 16 | Page-blocks-1-3-vs-4 | 472 | 10 | 15.85 | 1,7 | 10 | KEEL |
| 17 | Abalone-9-18 | 731 | 7 | 16.40 | 9 | 62 | KEEL |
| 18 | Glass5 | 214 | 9 | 22.77 | 7 | 60 | KEEL |
| 19 | Yeast-2-vs-8 | 482 | 8 | 23.10 | 4 | 71 | KEEL |
| 20 | GD3 | 5500 | 30 | 24.70 | 11 | 91 | Scikit learn |
| 21 | Yeast4 | 1484 | 8 | 28.10 | 6 | 57 | KEEL |
| 22 | Yeast5 | 1484 | 8 | 32.73 | 3 | 48 | KEEL |
| 23 | GD6 | 1600 | 25 | 38.00 | 7 | 88 | Scikit learn |
| 24 | Wineqlty-white-3-vs-7 | 900 | 11 | 44 | 4 | 63 | KEEL |
| 25 | Wineqlty-red-8_vs_6-7 | 855 | 11 | 46.5 | 5.7 | 67 | KEEL |
| 26 | GD1 | 2500 | 40 | 54.55 | 5.2 | 93 | Scikit learn |
| 27 | Wineqlty-white-3-9-vs-5 | 1482 | 11 | 58.28 | 3.77 | 60 | KEEL |
| 28 | GD4 | 2500 | 13 | 59.00 | 5 | 76 | Scikit learn |
| 29 | GD5 | 2000 | 5 | 59.61 | 4.55 | 55 | Scikit learn |
| 30 | GD7 | 6000 | 100 | 60.85 | 4.46 | 96 | Scikit learn |
| 31 | GD8 | 3500 | 15 | 62.63 | 4.7 | 83 | Scikit learn |
| 32 | GD9 | 10,000 | 50 | 62.69 | 4.7 | 94 | Scikit learn |
| 33 | Abalone-20-vs-8-9-10 | 1916 | 7 | 72.69 | 2.71 | 58 | KEEL |
| 34 | GD10 | 10,100 | 60 | 72.72 | 4 | 95 | Scikit learn |
| 35 | Poker-8-vs-6 | 1477 | 10 | 85.88 | 0.94 | 78 | KEEL |
| 36 | GD11 | 3000 | 3 | 99.00 | 3.3 | 61 | Scikit learn |
| 37 | Abalone19 | 4174 | 7 | 129.44 | 2 | 59 | KEEL |
| 38 | GD12 | 6000 | 20 | 156.89 | 2.2 | 90 | Scikit learn |
| 39 | GD13 | 5000 | 5 | 165.66 | 1.7 | 66 | Scikit learn |
| 40 | GD14 | 10,000 | 60 | 221.22 | 1.43 | 92 | Scikit learn |

## 6. Results

In this section, the values of the metrics used to evaluate the performance of the classifiers and the effectivity of the resampling strategies are reported first. Then, the results of the statistical significance tests conducted to compare the results and determine the best resampling method for every dataset. The last subsection includes the association rule models developed in order to find out the most suitable resampling method according to dataset characteristics.

### 6.1. Results from Individual Metrics

In this section, the performance of all resampling strategies for all quality metrics regarding dataset characteristics (number of attributes, number of instances, imbalance ratio, borderline examples and overlapping between classes) is analyzed. The results of the metrics are represented through radar charts where their values are placed on the radius of the circle and the characteristic under study on the external side of its perimeter. Original and resampled datasets are represented as lines into the circle in different colors.

In addition, statistical significance tests were conducted to compare the results and determine the best resampling method for every dataset. Finally, association rule models were developed in order to find out the most suitable resampling method according to dataset characteristics.

Accuracy is the first metric to be studied; it gives the percentage of instances correctly classified, but it does not provide information about which is the success rate in each class. Figure 1 shows the accuracy achieved by applying resampling techniques to different datasets and the impact of different data characteristics. The finding that can be appreciated at first sight is the absence or very little improvement achieved with resampling techniques regardless of the number of attributes, number of instances, imbalance ratio, borderline examples and overlapping classes. This fact is not surprising because random forest is an algorithm that produces good results in data imbalance contexts, achieving an overall acceptable accuracy although the accuracy of the minority class may be low. The most remarkable aspect is the drastic decrease of accuracy when using the RUS method. This can be due to the loss of information that the undersampling technique causes.

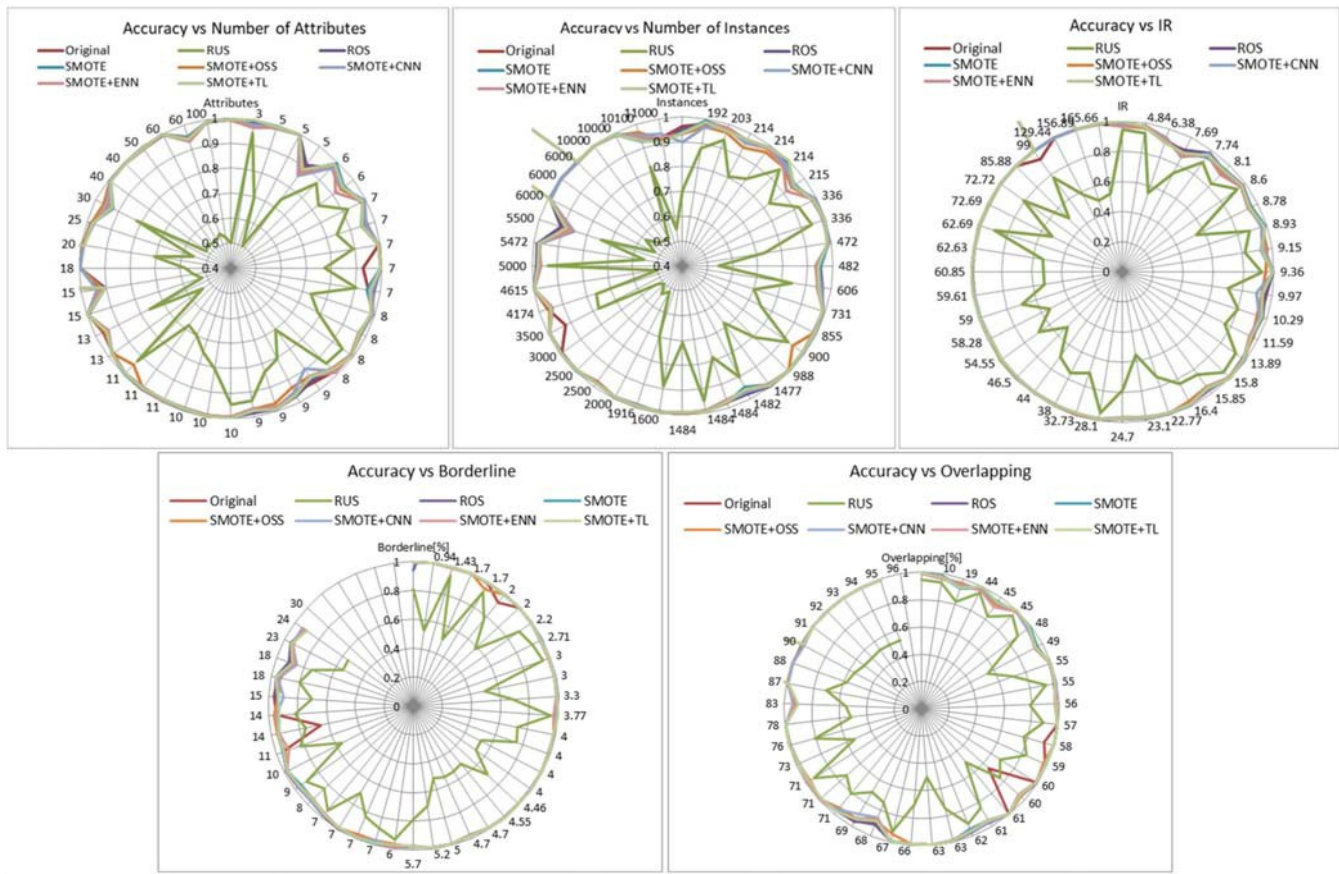

**Figure 1.** Influence of dataset characteristics on accuracy for all resampling strategies.

AUC is a metric that may be more useful than accuracy since it considers the rate of true positives and the rate of false positives. The closer AUC is to 1, the farther the classifier is from the random classification, so it will be better. Regarding this measure, given in Figure 2, resampling methods present an irregular behavior, and it can only be clearly appreciated that SMOTE + TL overcomes the others for high imbalance ratios (above 58) while RUS keeps giving bad results. We can also observe from the graphs the decrease of the AUC values as borderline examples and class overlapping increases, irrespective of the resampling strategy applied.

Figure 3 displays average precision obtained for minority and majority classes in the experiments. Unlike the previous metrics, RUS showed the best values in all cases while original datasets showed the worst in most of them. The rest of the resampling methods presented an irregular performance. High values of RUS can be explained by the fact that if there are few minority class predictions, the denominator of the precision formula is small and therefore the precision is high even when there are few correct minority class predictions. The improvement of the precision of RUS in case of borderline examples may be caused by the fact that the small sample size caused by RUS reduces noise rate. In the case of overlapping of classes, whereas the overlapping rate increases the precision decreases for all resampling strategies.

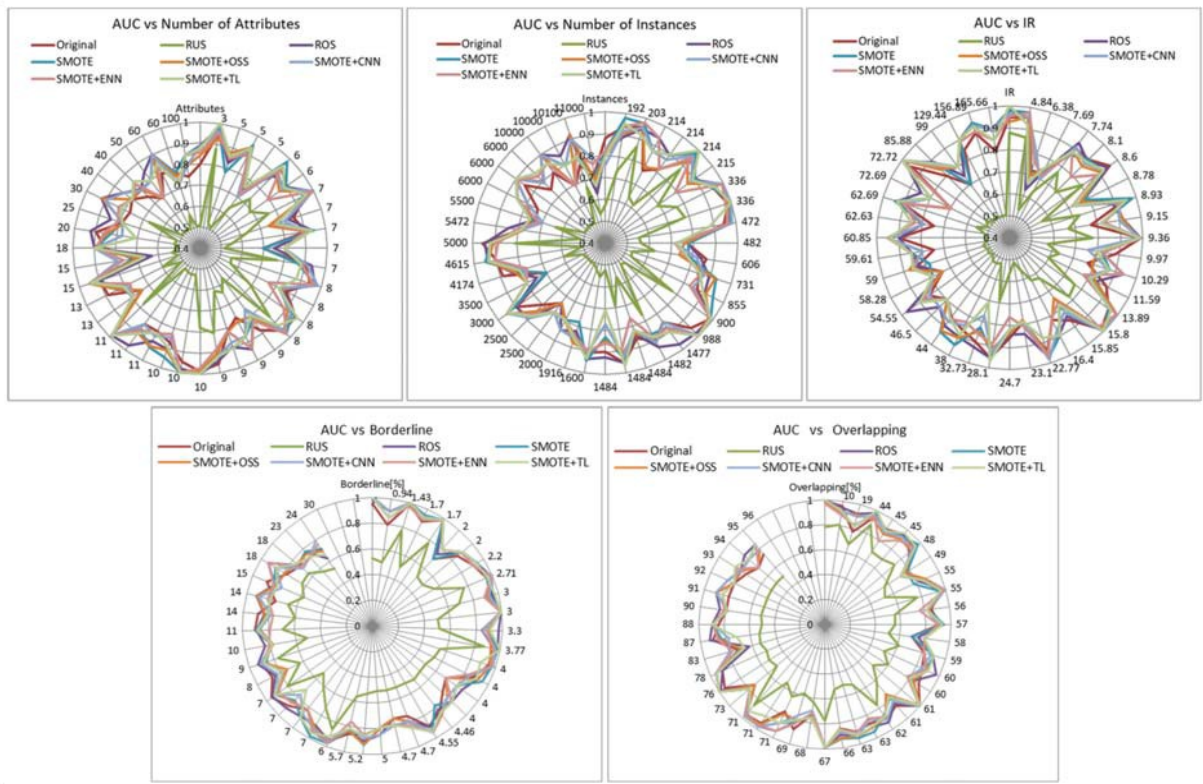

**Figure 2.** Influence of dataset characteristics on AUC for all resampling strategies.

For recall metric (Figure 4), similar conclusions to those obtained for AUC can be done. SMOTE + TL is the most suitable method for high imbalance ratio, followed by SMOTE, SMOTE + OOS and SMOTE + ENN, which gave very similar results to SMOTE + TL, except in some specific case in which they were worse. SMOTE + ENN is the most indicated for high percentage of borderline examples and for a small number of instances, and ROS for a medium number of instances. RUS method gave again the worst results.

F-Measure combines precision and recall since an increase in precision values often occurs at the expense of decreasing recall values. We have used the harmonic measure of precision and recall where both metrics have the same weight in the computation of F-Measure. Its results are shown in Figure 5. SMOTE + TL, and ROS gave the best results for high values of IR (greater than 58) and high overlapping classes, especially when borderline percent is low. SMOTE + OSS had a comparable performance for imbalance ratio greater than 60. It can also be observed that, in general, the F-measure decreases as the number of attributes, number of instances and overlap percentage increase.

The values of the next examined metric, G-Mean, are presented in Figure 6. The information provided by this metric is similar as the obtained with F-measure for the best methods, although here the poor performance of RUS is much more evident. Here again, there is a clear decrease in G-Mean for high overlapping values.

IBA is a metric that was proposed to overcome the problems of the before metrics by differentiating the contribution of the instances of each class to the accuracy. In Figure 7, showing its values, it can be observed that ROS and SMOTE + TL are the best for an imbalance ratio greater than 54 and in most of the cases of borderline examples. For class overlapping, we also observe that the performance decreases as the number of overlapping class increases. SMOTE + OSS achieved a similar performance for ratios greater than 59. Conclusions for the metric OP (Figure 8) were the same. In both cases RUS was the worst resampling technique.

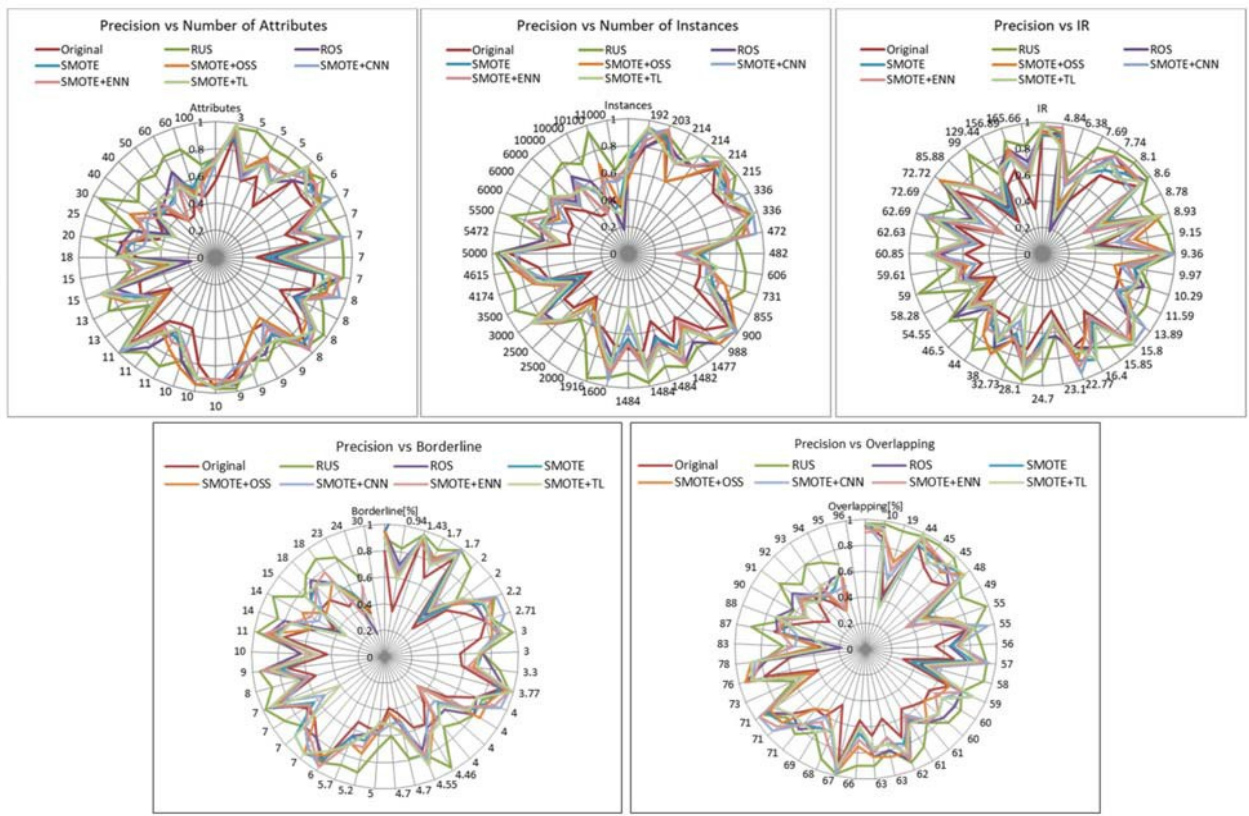

**Figure 3.** Influence of dataset characteristics on precision for all resampling strategies.

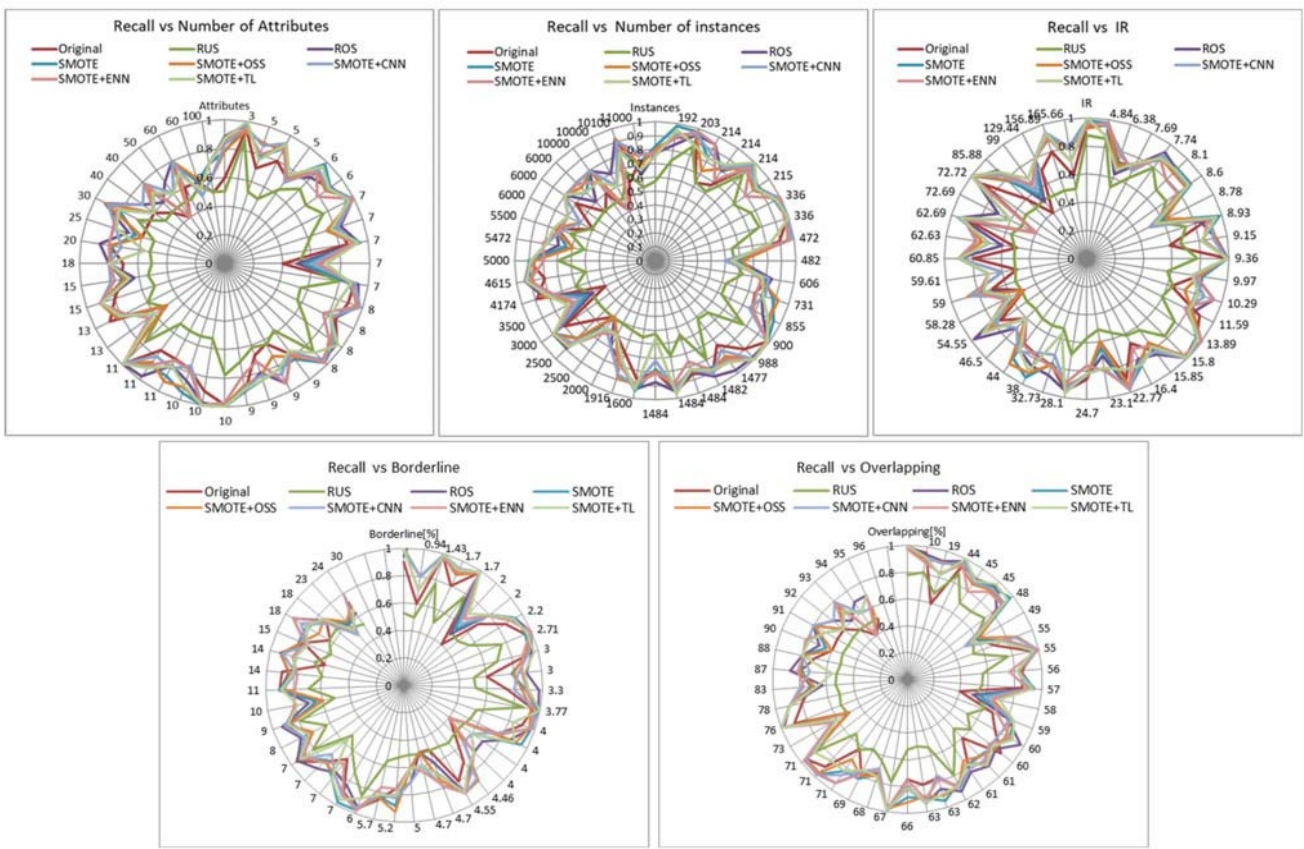

**Figure 4.** Influence of dataset characteristics on recall for all resampling strategies.

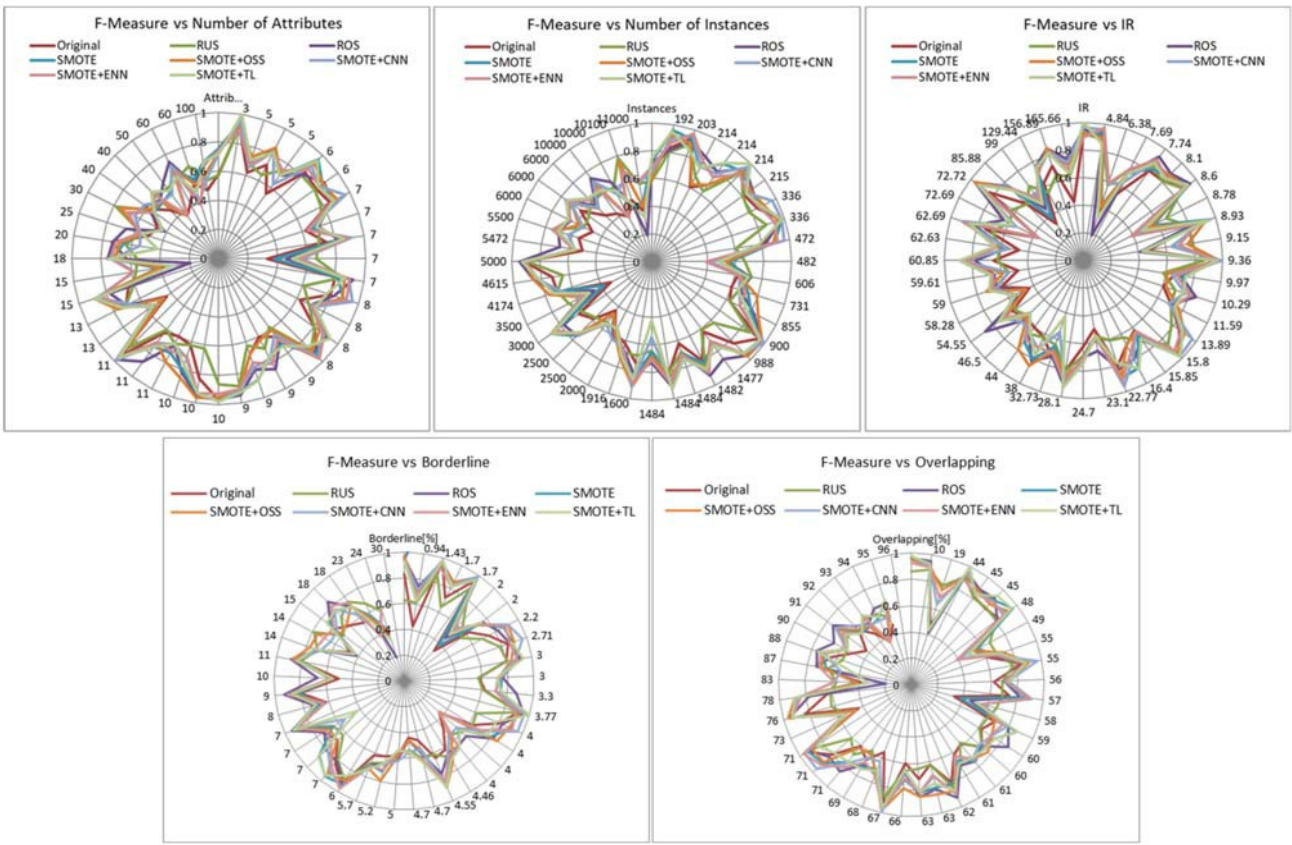

**Figure 5.** Influence of dataset characteristics on F-measure for all resampling strategies.

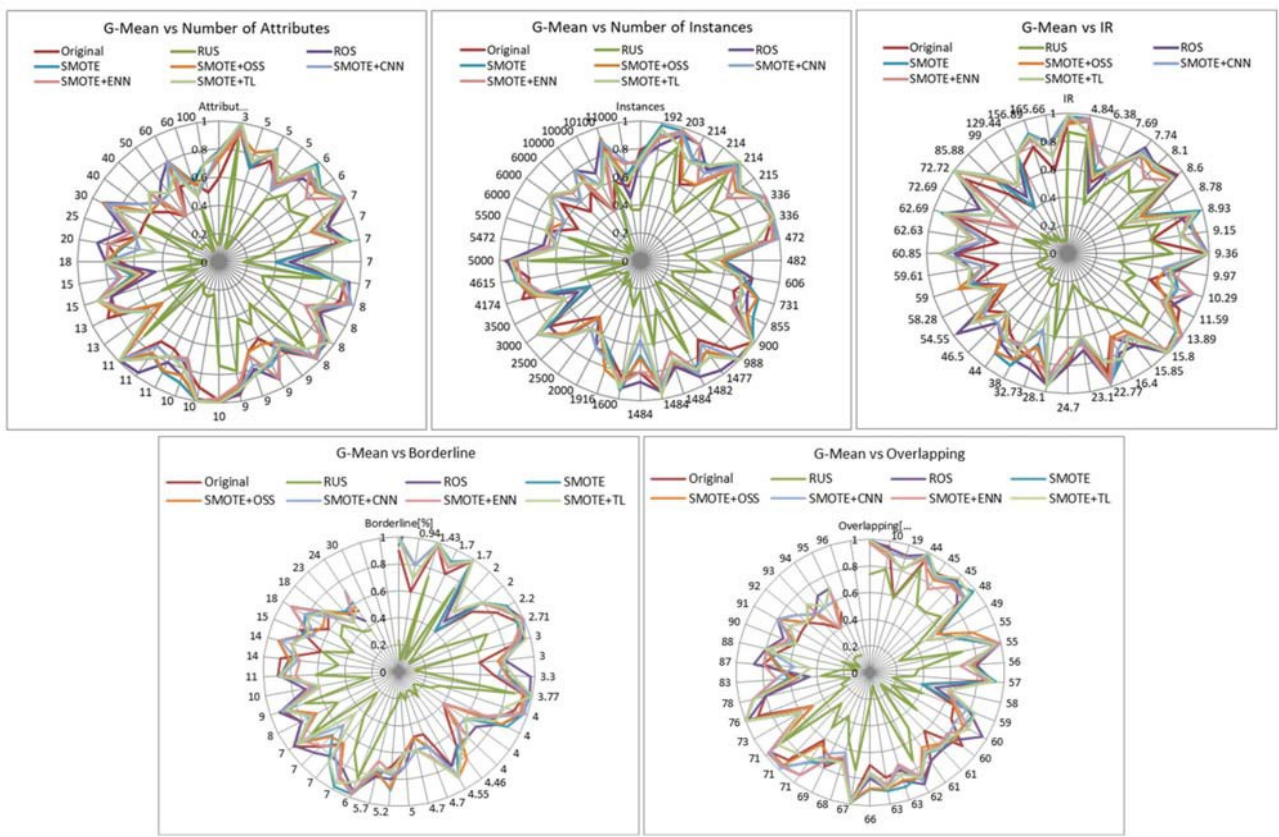

**Figure 6.** Influence of dataset characteristics on G-mean for all resampling strategies.

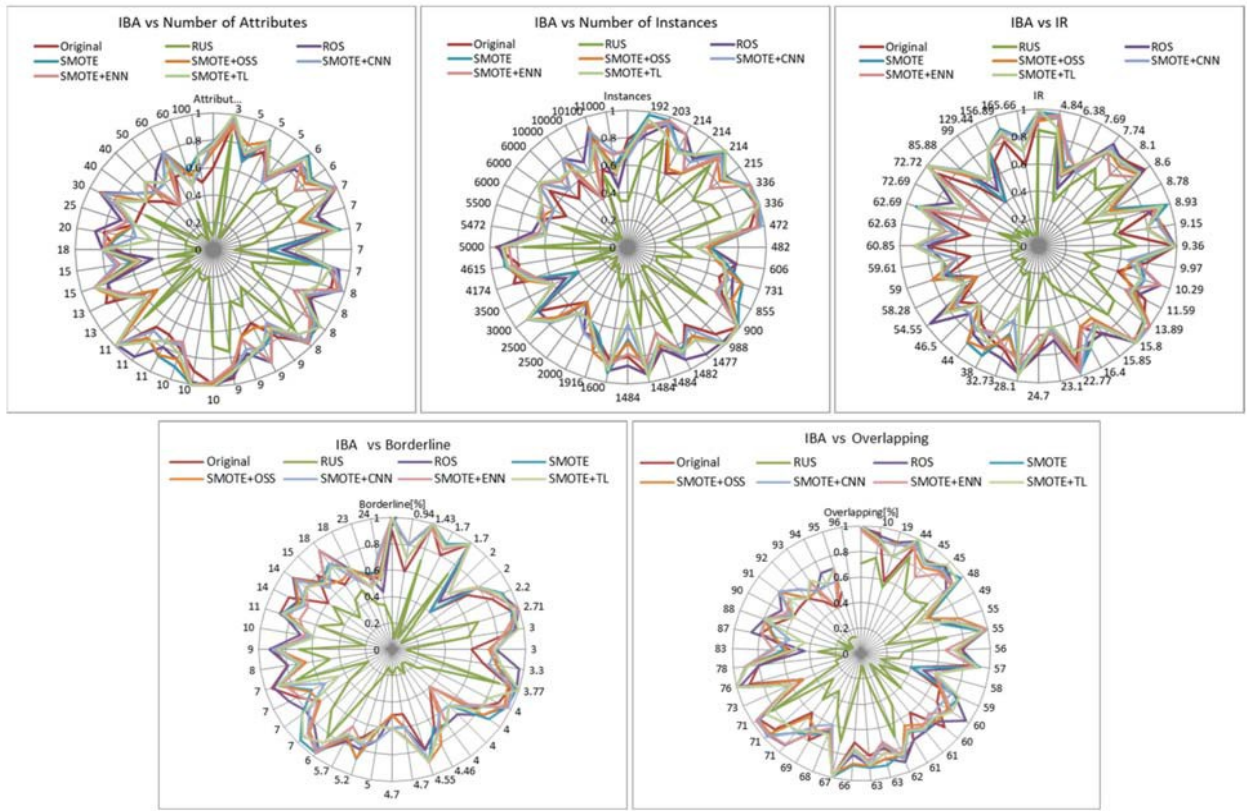

**Figure 7.** Influence of dataset characteristics on IBA for all resampling strategies.

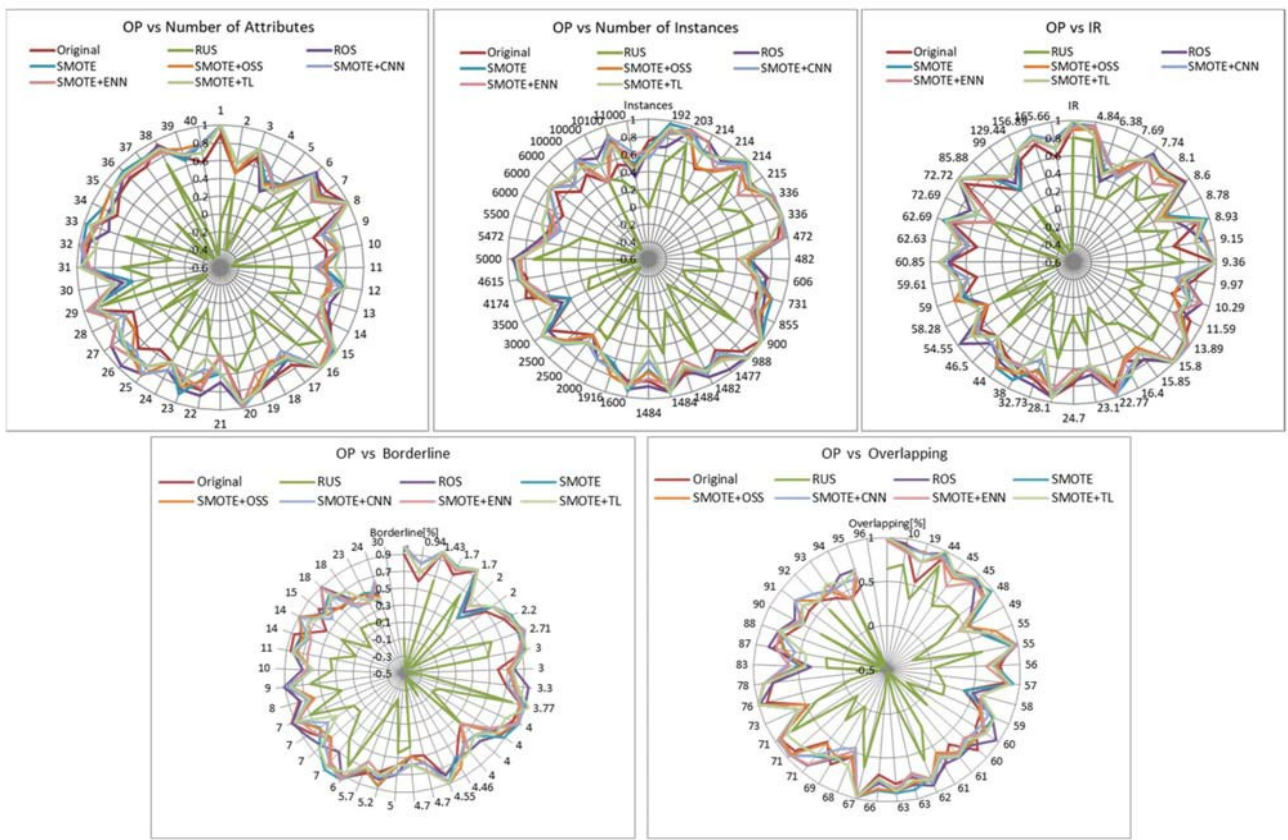

**Figure 8.** Influence of dataset characteristics on OP for all resampling strategies.

In view of these results, we can conclude that the last three metrics, specifically indicated for imbalanced classification problems, are similarly affected by the factors studied, with class overlap having the greatest impact.

*6.2. Statistical Significance Tests*

In order to reach reliable conclusions about the results reported above and to be able to induce the association rule models, it was necessary to find statistically significant differences among the results produced by different resampling methods. Thus, we performed Friedman and post hoc tests to determine the superiority of the strategy achieving the best result over the other evaluated resampling strategies for each dataset used in this study. The Friedman test is a popular and powerful procedure that allows to detect significant differences in the results of two or more algorithms for multiple performance comparisons [82]. Multiple comparisons involve multiple simultaneous statistical tests, each of them focused on and aspect of the problem. In this work, several performance metrics are evaluated for each dataset. In this way, the finding of the best resampling strategy for each dataset is reinforced since it is the result of the evaluation through multiple metrics. Figure 9 shows the graph plots of some datasets for the statistical comparison of the results, which present the order of the resampling methods in terms of all metrics. Numbers in the boxes represents the average ranks across all evaluation results for each resampling method tested in the study. The lower the rank the better the method is, and the greater the difference between the ranks is the more significant the difference between the methods being compared.

The best resampling strategies selected by these tests from the results of the performance metrics was then used to generate the association rules that relate the characteristics of the datasets to the most appropriate strategy for each of them.

*6.3. Association Rule Models for Selecting the Best Resampling Strategy*

After the qualitative analysis of each quality metric and after finding the best resampling method for every dataset, the next step was to perform a quantitative analysis to obtain a deeper knowledge about the impact of data properties on resampling strategies. For this purpose, we resorted to the induction of association rules that relate the characteristics of each dataset with the method that provided the best results regarding both individual metrics and globally. To do this, each dataset with the information of its characteristics was labeled with the resampling method that gave the best result for it, then the well-known Apriori algorithm for mining association rules was applied. In order to be able to apply this algorithm, the attributes were discretized into five equal-size intervals.

Since the number of examples from which the rules are generated is small, it has not been possible to generate class association rules. Therefore, the Apriori algorithm has been applied to obtain conventional association rules and those containing items corresponding to the classes have been selected. The quality measures of confidence, lift, leverage and conviction were used to evaluate the quality of the association rules.

Confidence is defined as the proportion of examples containing the consequent and the antecedent of the rule in relation to the proportion of examples containing the antecedent. It takes values between 0 and 1. Lift indicates the probability of the rule in relation to the joint probability of the antecedent and the consequent if they were independent. Rules are valid if lift values are greater than 1. Leverage measures the difference between the probability of the rule and the joint probability of the antecedent and the consequent if they were independent. It takes values in the interval [−1, 1]. Conviction assesses the degree to which the antecedent term influences the occurrence of the consequent term of an association rule. It takes values from 0 to infinite and the rule is only acceptable for values higher than 1.

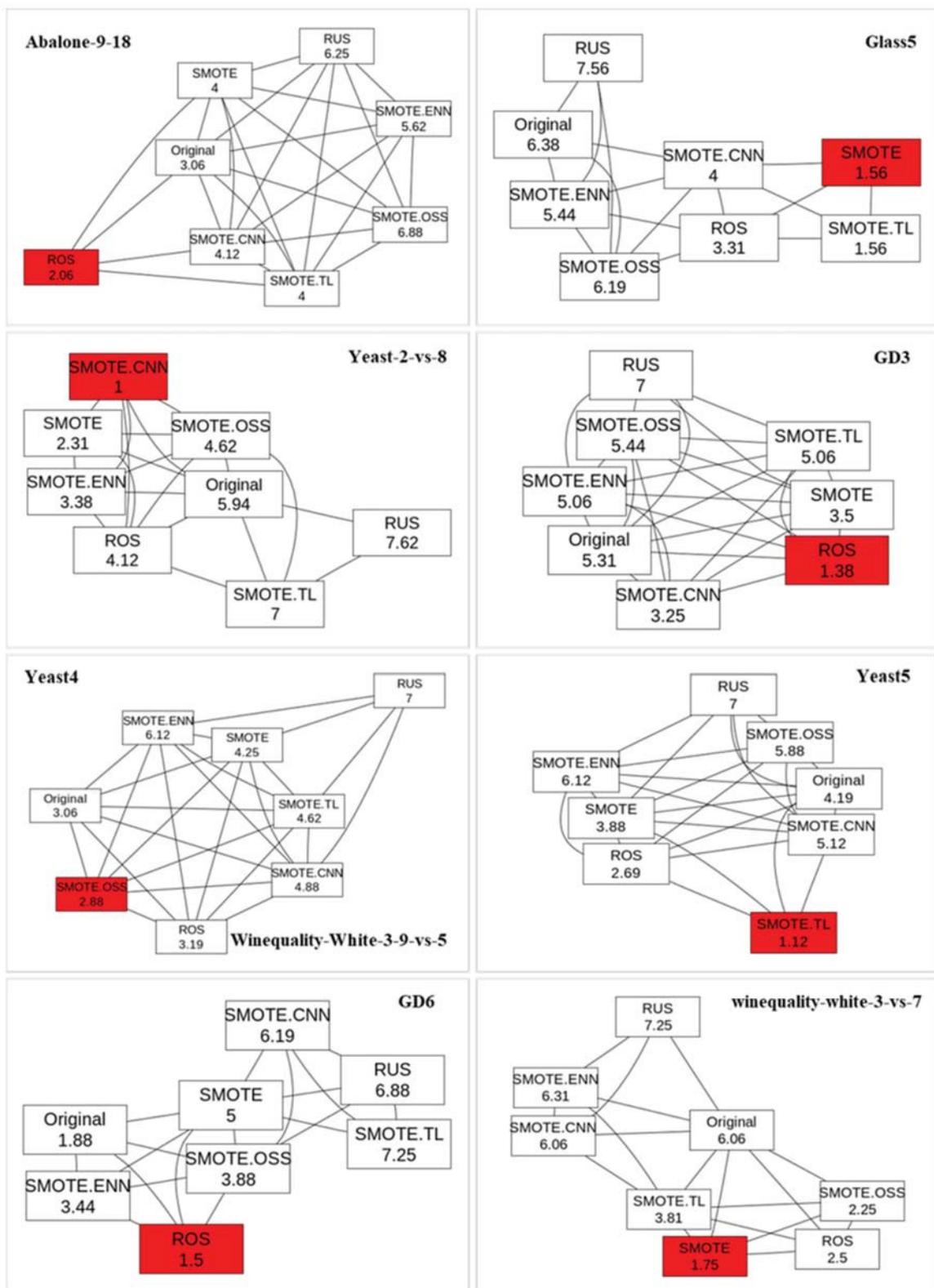

**Figure 9.** Graphs of the significance test results for some datasets.

The first association model was generated for individual metrics. Taking into account that the most specific metrics for evaluating imbalanced data classification are OP and IBA, we resorted to the induction of association rules that relate the characteristics of each dataset with the method that provided the best value of these metrics. The Apriori algorithm provided more than a hundred rules that exceeded the established confidence (0.9) and support (0.05) thresholds. The second model was induced taking into account the outcome of the significance tests performed considering all the metrics. Tables 2 and 3 present respectively the best rules of both models and the corresponding quality indicators: confidence (conf.), lift, leverage (lev.) and conviction (conv.). The complete models of association rules will be useful in selecting the best resampling strategy for any dataset once its characteristics are known.

Table 2. Evaluation of the best method regarding IBA values by means of association rules.

| Data Characteristics | Best IBA | Conf. | Lift | Lev. | Conv. |
| --- | --- | --- | --- | --- | --- |
| IR = (−∞–48] & BL% = (6.75–12.56] & OVL% = (78.8–∞) | ROS | 1.00 | 9.00 | 0.03 | 1.78 |
| IR = (−∞–48] & BL% = (12.56–18.38] & OVL% = (44.4–61.6] | Original | 1.00 | 6.75 | 0.03 | 1.70 |
| IR = (−∞–48] & BL% = (−∞–6.75] & OVL% = (44.4–61.6] | SMOTE + TL | 1.00 | 3.00 | 0.02 | 1.33 |
| BL% = (−∞–6.75] & (61.6–78.8] | SMOTE + CNN | 1.00 | 1.64 | 0.01 | 0.78 |
| IR = (48–91] & BL% = (−∞–6.75] & OVL% = (61.6–78.8] | SMOTE | 1.00 | 1.64 | 0.01 | 0.78 |
| IR = (48–91] & & BL% = (−∞–6.75] & OVL% = (78.8–∞) | ROS | 1.00 | 1.64 | 0.01 | 0.78 |
| IR = (−∞–48] & #Instances = (−∞–3154] | SMOTE | 1.00 | 1.54 | 0.05 | 2.46 |
| #Instances = (−∞–3007] & #Attributes = (−∞–22] & OVL% = (61.6–78.8] | SMOTE | 1.00 | 1.29 | 0.02 | 1.11 |
| #Instances = (−∞–3007] & OVL% = (44.4–61.6] | SMOTE | 1.00 | 1.54 | 0.03 | 1.41 |
| #Attributes = (−∞–22] & BL% = (−∞–6.75] | SMOTE + OSS | 1.00 | 1.29 | 0.02 | 1.11 |

Table 3. Evaluation of the best method by means of association rules.

| Data Characteristics | Best Method | Conf. | Lift | Lev. | Conv. |
| --- | --- | --- | --- | --- | --- |
| IR = (−∞–48] & BL% = (12.56–18.38] | Original | 1.00 | 8.00 | 0.04 | 1.75 |
| IR = (−∞–48] & BL% = (6.75–12.56] & OVL% = (78.8–∞) | ROS | 1.00 | 5.00 | 0.04 | 1.60 |
| BL% = (6.75–12.56] & OVL% = (44.4–61.6] | SMOTE | 1.00 | 4.44 | 0.04 | 1.55 |
| IR = (−∞–48] & BL% = (−∞–6.75] & OVL% = (44.4–61.6] | SMOTE + TL | 1.00 | 4.44 | 0.04 | 1.55 |
| IR = (48–91] & OVL% = (78.8–∞) | SMOTE + TL | 1.00 | 4.00 | 0.04 | 1.50 |
| IR = (134–178] | ROS | 1.00 | 2.86 | 0.03 | 1.30 |
| BL% = (6.75–12.56] & OVL% = (78.8–∞) | ROS | 1.00 | 2.86 | 0.03 | 1.30 |
| #Instances = (−∞–3007] & OVL% = (44.4–61.6] | SMOTE | 1.00 | 1.48 | 0.04 | 1.63 |
| #Attributes = (−∞–22] & BL% = (−∞–6.75] | SMOTE + CNN | 1.00 | 1.29 | 0.02 | 0.68 |
| IR = (48–91] & #Attributes = (−∞–22] | SMOTE + OSS | 1.00 | 1.29 | 0.02 | 0.68 |

## 7. Discussion

The individual results of the evaluation measures performed have allowed us to draw some preliminary conclusions on the impact of some data-related factors on the effectiveness of the resampling strategies. An important observation is the bad behavior of RUS that led to a worsening of all metrics except precision compared to those from original datasets, being more pronounced for high imbalance ratios. This is consistent with findings of other studies [24] that relate class balancing with RUS to the small sample size problem in presence of high imbalance ratio, which has a negative impact on the model performance. In contrast to RUS, the rest of the resampling strategies did manage to improve the results of all metrics obtained from the original datasets in most cases, regardless of the factors studied, as stated in the literature [14,79]. Furthermore, it has been found that the factors that most influence the loss of reliability of the classifiers according to the G-Mean, OP and IBA metrics, both with the original data and with the resampled data, are the high degree of overlap between classes and the number of instances. This behavior has also shown that these metrics are suitable for assessing the quality of the classification of unbalanced data, since class overlapping is one of the most influential factors in the precision of the minority class [14,73,74]. Other more specific findings related to the behavior of resampling strategies as a function of data characteristics, detailed in the results section, could not be compared with work in the literature because we have not found studies covering this range of techniques, metrics, datasets and related factors.

On the other hand, the association analysis, in addition to helping us to select the most effective resampling method according to the characteristics of the data, has allowed us to draw some interesting conclusions from the patterns discovered. From the first model, we deduce that the best IBA value is given by the original dataset, without resampling, when IR is very low, regardless of the other properties of the datasets, except in the case that the overlapping between classes is very high, then ROS is the best strategy. It can also be observed that SMOTE used alone or combined with CNN, OOS or TL is the strategy that obtains the best results in most of the rules. Nevertheless, SMOTE + ENN does not reach the best IBA value under any circumstances. In addition, we can mention other representative patterns: If the number of instances is very low, SMOTE is suitable when IR is medium-high while SMOTE + TL is indicated when IR is very low. The association rules also confirm the observation that RUS is the worst strategy since it does not appear in any of the rules of this model.

The conclusions from the second rule model are not as clear as in the case of IBA because the effectiveness of resampling strategies varies substantially depending on the metrics being analyzed. Still, some of the conclusions discussed in the case of IBA are applicable here as well. For example, the fact that RUS and SMOTE + ENN do not appear in the rule model as the best methods or that for low IR, the best results are obtained with the original dataset.

Finally, to reinforce the validity of the findings and conclusions obtained from this study, despite having worked with the same resampling rates for all datasets, we have analyzed the effect of the resampling methods on the most influential factors on the classification results. Previously in this section, we have pointed out that our work as well as others in the literature have proven that the overlap between classes and the number of instances have the highest impact on the classification of imbalanced data. Regarding the former, the values of class overlapping in the original datasets and in the datasets resampled by means of the different strategies have been plotted against the initial imbalance ratio. By using the fixed resampling rate for all datasets, the imbalance degree after resampling for the datasets with lower original IR will be lower than for the datasets with higher IR. However, as can be seen in Figure 10, this has no direct correlation on the magnitude of the change in class overlapping. Regarding the number of instances, related to the small sample size problem, the values of this factor after resampling depend mainly on the initial size of each dataset. From these observations, we can deduce that the degree of imbalance post resampling influences neither the class overlap nor the number of instances, and therefore would not greatly influence the final result of the classification. From this fact we can deduce that keeping the imbalance rates constant would not introduce a significant bias in the conclusions of the study.

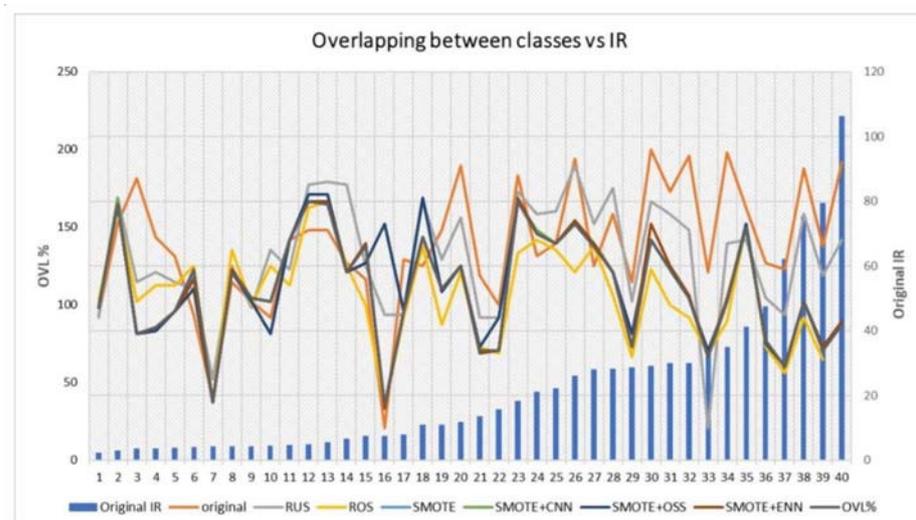

**Figure 10.** Effect of resampling methods on class overlapping for different values of original IR.

Additionally, we have tested the effects of applying SMOTE, which is the best performing method, with different resampling rates. In Figure 11, we show the results of accuracy, TPR, TNR, OP, IBA and G-mean obtained for four datasets representative of different IR values. As can be seen, 500% resampling provides the best results of all metrics except for very highly balanced datasets such as abalone9, which has an IR of 129.44. In these cases, increasing the resampling percentage may lead to an overfitting problem.

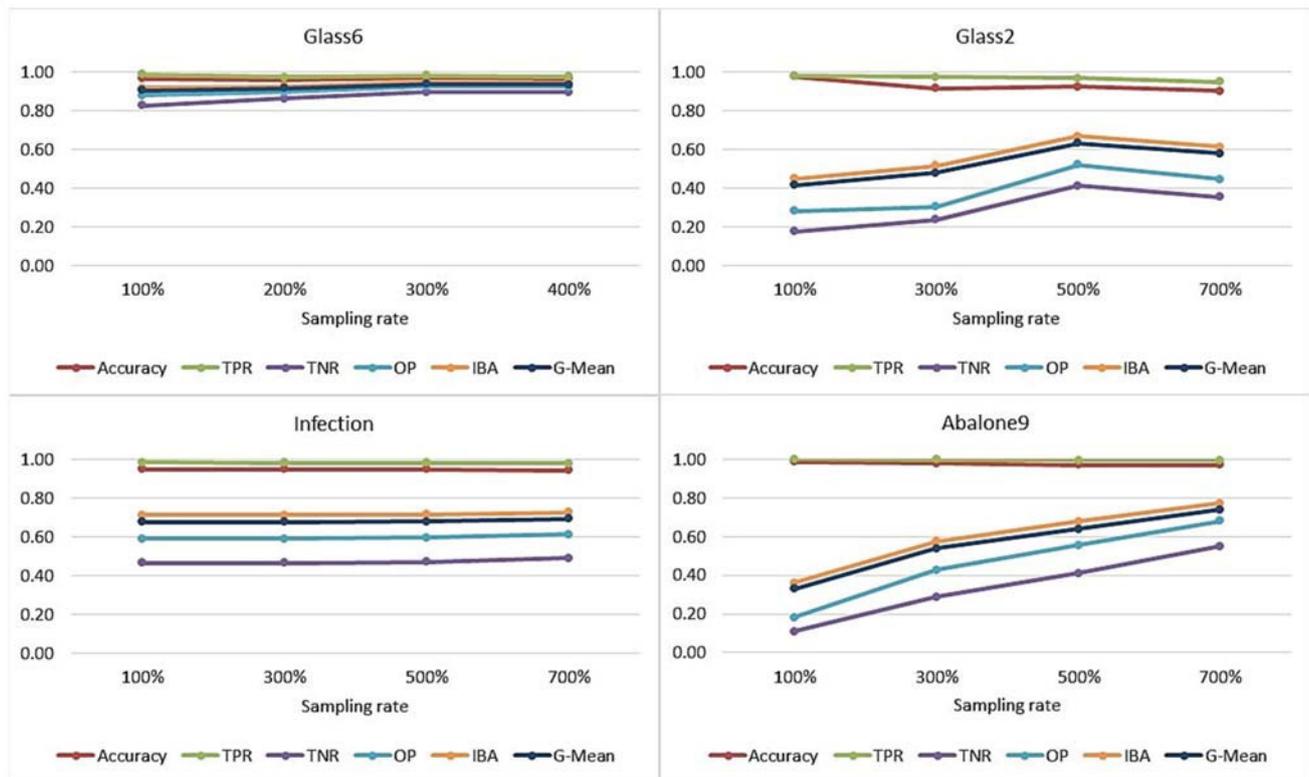

**Figure 11.** Effect of SMOTE resampling rate on accuracy, TPR, TNR, OP, IBA and G-mean with datasets representative of different IR.

## 8. Conclusions

The purpose of this work has been to provide some indications to help choosing the suitable resampling strategy for dealing with imbalanced data classification regarding some data properties. An extensive empirical study has been conducted in which 40 datasets covering a wide range of applications domains have been used to induce classification models both in their original form and resampled by means of seven different resampling methods. Data properties vary from 4.84 to 221.22 IR, from 192 to 15,000 instances, from 3 to 100 attributes, from 0.94 to 30 borderline example rate and from 10 to 96 overlapping class rate. As far as we know, no study covering such a range of these properties has been presented before.

Based on the results of previous work, all experiments were performed with random forest. Performance was evaluated through a total of eight metrics of both classic and new measures designed for the specific problem of imbalanced data. The values of these metrics and the results of the statistical significance tests carried out revealed some behaviors which were used as feed to create models of association rules that allowed us to draw more in-depth conclusions. These models of high-confidence rules involve useful patterns for selecting the best resampling strategy for each dataset. Despite the extensive study, the work has some limitations that will be addressed in the future. Among them is the fact that only the best-known resampling techniques with a fixed sampling rate are evaluated using a single classification algorithm. In addition, the association rule model would have been more reliable with results obtained from a much larger number of datasets. A more ambitious research direction for the future is the development of automatic methods for classification of imbalanced data that go beyond the simple selection of resampling techniques.

**Author Contributions:** Conceptualization, M.S.K., M.N.M.-G. and F.S.-H.; methodology, M.S.K., M.N.M.-G. and F.S.-H.; validation, M.S.K., M.N.M.-G. and F.S.-H.; investigation, M.S.K., M.N.M.-G. and F.S.-H.; data curation, M.S.K.; writing—original draft preparation, M.S.K.; writing—review and editing, M.N.M.-G. and F.S.-H.; supervision, M.N.M.-G. and F.S.-H.; project administration, M.N.M.-G.; funding acquisition, M.N.M.-G. All authors have read and agreed to the published version of the manuscript.

**Funding:** This work has been funded by the Junta de Castilla y León, Spain, grant number SA064G19.

**Institutional Review Board Statement:** Not applicable.

**Informed Consent Statement:** Not applicable.

**Data Availability Statement:** The datasets used in this study were obtained from public repositories such as KEEL (https://sci2s.ugr.es/keel, last accessed: 18 August 2021) or UCI (https://archive.ics.uci.edu/ml/, last accessed: 18 August 2021). Detailed information on each dataset and its origin is provided in Table 1.

**Conflicts of Interest:** The authors declare no conflict of interest.